\documentclass[]{ceurart}

\usepackage{listings}
\usepackage{booktabs}
\usepackage{multirow}
\usepackage{graphicx}
\usepackage{amsmath}
\usepackage{subcaption}
\usepackage[ruled]{algorithm2e}
\usepackage[justification=raggedright,singlelinecheck=false]{caption}

\begin{document}

\copyrightyear{2026}
\copyrightclause{Copyright for this paper by its authors.
  Use permitted under Creative Commons License Attribution 4.0
  International (CC BY 4.0).}

\conference{CLEF 2026 Working Notes, 21 -- 24 September 2026, Jena, Germany}

\title{Adversarial Deepfake Generation and an Investigation of
Purification-Based Adversarial Detection}

\title[mode=sub]{Notebook for the ImageCLEF at CLEF 2026}

\author[1]{Junghyun Kim}[%
  email=jhyeun1234@sch.ac.kr,
]
\cormark[1]
\author[1]{Seunghyun Kim}[%
  email=ksh011018@sch.ac.kr,
]
\author[1]{Jiyoung Woo}[%
  email=jywoo@sch.ac.kr,
]
\address[1]{Department of AI and Big Data Engineering, Soonchunhyang University, Asan, South Korea}
\cortext[1]{Corresponding author.}

\begin{abstract}
This paper describes the participation of team ``Go To Germany'' in
the ImageCLEF 2026 Deepfake Detection and Generation Task~\cite{ImageCLEFDeepfakeTaskOverview2026}.
For the image generation task, we employ FLUX.1-dev with PuLID for
identity-preserving face synthesis, combined with a multi-model PGD
adversarial attack targeting 12 detectors simultaneously
(DiffJPEG-in-loop, MI/DI/EoT, adaptive weighting, two-stage
warm-start). Our approach achieved $90\%$ evasion against organizer
detectors and $57.6\%$ against participant detectors, with a final
generation score of $0.4170$. For the image detection task, we
combine two complementary detectors --- SigLIP+DINOv2 for
AI-generated images and GenD-DINOv3~\cite{GenD2026} for face manipulations --- in
a max-probability ensemble, achieving $99.4\%$ accuracy on baseline
deepfakes but suffering from high false-positive rates on real
images, resulting in a final detection score of $0.6986$.
Beyond the official submission, we conducted a self-initiated
investigation of purification-based adversarial detection,
comparing three families of detection signals across six detectors
that share a CLIP ViT-L/14 backbone. We find that raw
$|\Delta\text{logit}|$ under median-$3$ purification, applied
through the EFFORT detector~\cite{EFFORT2025}, separates adversarial inputs from
clean inputs with AUROC $0.81$--$0.98$ across four adversarial
source types --- a finding that refutes the simple
backbone-preservation hypothesis and exposes a sharp JPEG-quality
cliff at $Q70$ where the signal collapses.
\end{abstract}


\maketitle

\section{Introduction}
\label{sec:introduction}

Deepfake technology has advanced to the point where synthetic media
can convincingly deceive both human observers and automated detection
systems. The ImageCLEF 2026 Deepfake Detection and Generation
Task~\cite{ImageCLEFDeepfakeTaskOverview2026}, part of the broader
ImageCLEF 2026 evaluation campaign~\cite{ImageCLEF2026}, challenges
participants to both generate realistic deepfakes that evade
state-of-the-art detectors and build robust detection systems capable
of identifying diverse manipulation types under real-world conditions.

Our team participated in both the image generation and image detection
sub-tasks. Observing during the generation track that adversarial
attacks can substantially degrade detection performance, we conducted
a self-initiated investigation of purification-based methods for
detecting adversarial perturbations, reported here as further
experiments beyond the competition submission.

Our main contributions are as follows:
\begin{itemize}
\item A multi-stage adversarial generation pipeline combining
FLUX.1-dev and PuLID with a custom PGD attack that simultaneously
targets 12 detectors and integrates DiffJPEG-in-loop, adaptive
per-image weighting, and two-stage warm-start beyond standard
PGD/MI/DI/EoT, achieving $\geq 95\%$ evasion on 15 of 17 evaluated
detectors (Section~\ref{sec:generation}).
\item A two-detector max-probability ensemble combining SigLIP+DINOv2
(AI-generated image specialist) and GenD-DINOv3~\cite{GenD2026} (face manipulation
specialist) for complementary deepfake detection, achieving $99.4\%$
accuracy on baseline deepfakes (Section~\ref{sec:detection}).
\item A further-experiments investigation of purification-based
adversarial detection across six detectors sharing a CLIP ViT-L/14
backbone, identifying EFFORT~\cite{EFFORT2025} (SVD-residual fine-tuning) as the only configuration that broadly generalizes (AUROC $0.81$--$0.98$ across
four adversarial source types in the raw input regime) and refuting
the simple backbone-preservation hypothesis
(Section~\ref{sec:purification}).
\end{itemize}

All experiments in this work were conducted on a server with
four NVIDIA~A100 $40$\,GB GPUs.


\section{Related Work}
\label{sec:related}

\subsection{Deepfake Generation}
\label{subsec:rw-generation}

Modern face manipulation methods fall into three categories by
mechanism. \emph{Face-swap} methods
(FaceShifter~\cite{FaceShifter2020}, SimSwap~\cite{SimSwap2020})
transfer identity onto a target face while preserving expression
and background. \emph{Face reenactment}
(Face2Face~\cite{Face2Face2016},
NeuralTextures~\cite{NeuralTextures2019}) preserves identity while
transferring expressions and pose. \emph{Lip synchronization} methods (Wav2Lip~\cite{Wav2Lip2020})
drive the mouth region alone. All three modify only local regions
of the frame, making detection challenging. A separate line,
\emph{full-face synthesis}, generates entire faces from scratch,
driven by an evolution from generative adversarial networks through
denoising diffusion models~\cite{DDPM2020}, Latent
Diffusion~\cite{LDM2022}, and the Diffusion Transformer (DiT)
architecture~\cite{DiT2023} that replaces the U-Net backbone with a
transformer operating on tokenized noised latent patches. Several
large-scale DiT systems have been released as this paradigm
matured --- including Stable Diffusion~3 and the FLUX family ---
of which we adopt the open-source FLUX.1-dev~\cite{FLUX2024} for
the compatibility of its ecosystem with the ControlNet and
identity-injection modules described below.

For identity-preserving generation, base diffusion generators are
augmented with two families of conditioning modules. Structural
controllers such as ControlNet~\cite{ControlNet2023} and its unified
variants~\cite{ControlNetUnion2024} enforce geometric constraints,
typically using the OpenPose~\cite{OpenPose2019} convention.
Identity-injection modules --- IP-Adapter~\cite{IPAdapter2023},
PuLID~\cite{PuLID2024} --- add subject-specific features through
cross-attention or contrastive alignment without per-identity
fine-tuning. The off-the-shelf assembly of DiT, structural
ControlNet, and identity-injection module underlies the generation
pipeline of Section~\ref{sec:generation}.

\subsection{Deepfake Detection}
\label{subsec:rw-detection}

Deepfake detection has evolved rapidly, driven by an arms race with
improving generation methods. Early efforts focused on hand-crafted
artefacts \emph{specific to face manipulation}: physiological
inconsistencies, blending-region signatures (Face
X-ray~\cite{FaceXray2020}), and frequency-domain irregularities
specific to GAN upsampling~\cite{Frank2020,Durall2020}.
Data-synthesis strategies such as Self-Blended
Images~\cite{SBI2022} extended this line by generating pseudo-fakes
at training time. These signal-level cues were designed for the
face-swap and reenactment regime dominant at the time and degrade
quickly under real-world post-processing~\cite{DeepfakeBench2023}.

A second wave targeted cross-generator generalization while staying
within the face-manipulation regime: latent-space
augmentation~\cite{LSDA2024} perturbs intermediate representations
during training, and dual data alignment~\cite{DDA2025} aligns
train-test distributions. Complementary temporal-modeling approaches
--- LipForensics~\cite{LipForensics2021}, FTCN~\cite{FTCN2021} ---
exploit frame-level inconsistencies that image-level detectors miss.
These methods typically operate on face-cropped inputs and target
the manipulation types collected in FaceForensics++.

The most recent shift adopts representation-based detection built
on large pretrained vision foundation models and, critically,
\emph{broadens scope from face-only manipulation to a unified
regime that also covers full-image AI synthesis}. Ojha et
al.~\cite{UnivFD2023} first demonstrated that a linear classifier
on frozen CLIP~\cite{CLIP2021} features generalizes across
synthetic-image generator families, motivating parameter-efficient
specialization strategies for VFM backbones --- CLIP,
SigLIP~\cite{SigLIP2023}, and self-supervised vision transformers
DINOv2~\cite{DINOv2_2023} and DINOv3~\cite{DINOv3_2025} --- such as
LoRA low-rank tuning~\cite{LoRA2022,KongLoRAForgery2023}, adapter
modules in ForAda~\cite{ForAda2025}, LayerNorm-only fine-tuning in
GenD~\cite{GenD2026}, and SVD-residual decomposition in
EFFORT~\cite{EFFORT2025}. This paradigm shift toward unified
detection is exemplified by UNITE~\cite{UNITE2025}, which builds on
a SigLIP backbone with an attention-diversity loss to detect both
face-manipulation and fully AI-generated content within a single
model without requiring face-crop preprocessing. Very recent work
on frozen DINOv3 features~\cite{FGTS2025} further argues that
foundation-model representations already encode global,
low-frequency structural cues transferable across generator
families, though with caveats about localized face editing where
signature preservation remains challenging.

Two observations from this literature inform our design. First,
even in the emerging era of unified detectors, face manipulation
produces \emph{local} artefacts around the modified region, while
full-image AI synthesis produces \emph{global} artefacts spanning
the image; on our internal benchmark
(Section~\ref{subsec:det-model-select}) these signatures remain
qualitatively different enough that an ensemble of specialists
outperforms any single unified detector we surveyed, motivating the
two-detector architecture of Section~\ref{sec:detection}. Second,
the diversity of parameter-efficient fine-tuning strategies on a
shared backbone raises the question of whether they differ in
properties beyond raw classification accuracy --- specifically,
their behavior under input perturbations relevant to adversarial
detection --- which forms the basis of our
Section~\ref{sec:purification} investigation.

\subsection{Adversarial Robustness of Deepfake Detectors}
\label{subsec:rw-adversarial}

Deep neural network classifiers, including deepfake detectors,
exhibit a fundamental vulnerability: their predictions can be
reversed by \emph{adversarial perturbations} --- small, carefully
crafted modifications to an input, imperceptible to human observers,
that change the classifier's output class. In the deepfake context
this creates a concrete threat model: an attacker can post-process
a fake image with a tailored noise pattern that drives the
detector's fake-class probability below the decision threshold,
effectively laundering the deepfake past automated detection.
Perturbations are typically constrained to small $\ell_\infty$
norm (each pixel modified by at most $\epsilon$, commonly
$\epsilon = 2/255$ or $8/255$). The standard attack family
progresses from single-step FGSM~\cite{FGSM2015} through multi-step
BIM~\cite{BIM2017} to projected gradient descent
(PGD)~\cite{PGD2018}, with Carlini \& Wagner~\cite{CW2017}
representing an optimization-based alternative and
AutoAttack~\cite{AutoAttack2020} providing an ensemble benchmark
that resists gradient-masking artefacts. Transferability
techniques --- Momentum Iterative~\cite{MomentumIterative2018},
Input Diversity~\cite{InputDiversity2019}, Expectation over
Transformation~\cite{EoT2018} --- boost cross-model generalization,
and DiffJPEG~\cite{DiffJPEG2017} provides a differentiable
surrogate for non-differentiable post-processing.

Defenses fall into three broad camps. Adversarial
training~\cite{PGD2018} jointly trains on clean and perturbed
inputs at substantial compute cost and typically at the price of
clean accuracy. Input purification transforms the input before
classification to remove perturbations while preserving semantic
content; Feature Squeezing~\cite{FeatureSqueezing2018} is the
canonical instantiation and additionally proposes a detection
variant that flags an input as adversarial when the classifier's
predictions disagree between the original and squeezed views.
Detection-based defenses separately flag adversarial inputs
without modifying the classifier --- the Mahalanobis-distance
framework of Lee et al.~\cite{Mahalanobis2018}, developed and
evaluated on generic out-of-distribution benchmarks such as
CIFAR-10/100 and SVHN, treats them as off-manifold samples
detectable by per-class Gaussian fits on intermediate features,
with complementary lines using local intrinsic
dimensionality~\cite{LID2018}. Both purification-based detection
and feature-space outlier detection have been studied extensively
on generic image classifiers, but their behavior on modern
foundation-model-based deepfake detectors --- where the choice of
parameter-efficient fine-tuning strategy varies substantially
across detectors sharing a common backbone --- has not been
systematically characterized. Our Section~\ref{sec:purification}
investigation addresses this gap by quantifying the output logit
shift under a median-filter operator across six detectors that
share a CLIP ViT-L/14 backbone but differ in fine-tuning
strategy, isolating fine-tuning method as the variable that
governs signal strength.

\subsection{Datasets}
\label{subsec:rw-datasets}

FaceForensics++~\cite{FF2019} is the de-facto benchmark for face
manipulation, providing five manipulation methods (Deepfakes,
Face2Face~\cite{Face2Face2016}, FaceSwap,
FaceShifter~\cite{FaceShifter2020},
NeuralTextures~\cite{NeuralTextures2019}) over 1{,}000 source
videos at three compression levels; larger and more diverse
successors include Celeb-DF~\cite{CelebDF2020},
DFDC~\cite{DFDC2020}, and DF40~\cite{DF40_2024}. For the
full-image AI-synthesis regime distinct from face manipulation,
GenImage~\cite{GenImage2023} and
DiffusionForensics~\cite{DIRE2023} are standard cross-generator
benchmarks, with DiFF~\cite{DiFF2024} targeting
diffusion-generated faces specifically; because these benchmarks do not cover the FLUX-family generators
we expected to appear both in our own submissions and in
participant deepfakes, we additionally use FLUX-inclusive public
pools such as BitMind on HuggingFace for detector model selection
and calibration. The ImageCLEF 2026 Deepfake Detection
task~\cite{ImageCLEFDeepfakeTaskOverview2026} we participate in
combines organizer-generated face-manipulation deepfakes,
participant-submitted AI-generated images, and curated real
images into a heterogeneous evaluation scored under a weighted
rule that emphasizes cross-team generalization.

\section{Image Generation}
\label{sec:generation}

\subsection{Task Overview}

The generation task requires producing 1{,}002 deepfake face images
($256{\times}256$ PNG) corresponding to 335 target identities. The
organizers provide one reference video per identity (12--45 seconds
of a single person speaking under varied lighting and pose) and
1--3 sets of MediaPipe FaceMesh landmark coordinates per identity
that serve as facial geometry constraints for generation. The
evaluation criteria are: (1) landmark distance to the reference
geometry, (2) identity similarity via face recognition embeddings,
and (3) deepfake evasion scores against both organizer and
participant detection systems, with evasion weighted most heavily
in the final score.

\subsection{Base Generation Pipeline}
\label{subsec:base-generation}
The base image synthesis assembles three off-the-shelf components without
architectural modification (Figure~\ref{fig:base-pipeline}):
\begin{enumerate}
    \item \textbf{FLUX.1-dev}~\cite{FLUX2024}: a Diffusion Transformer
    generates 512$\times$512 face images using a fixed prompt describing
    a person against a light blue wall with natural lighting.
    \item \textbf{ControlNet-Union-Pro-2.0}~\cite{ControlNetUnion2024}:
    enforces facial geometry and body pose constraints via its
    pose conditioning mode. Because this ControlNet variant is
    pretrained on the OpenPose~\cite{OpenPose2019} input convention
    rather than the MediaPipe convention used by the competition's
    supplied landmarks, we render the provided MediaPipe 478-point
    face landmarks and 33-point body pose into an OpenPose-style
    control image: the 33 MediaPipe body joints are mapped to the 18
    OpenPose keypoints with the standard colored-limb encoding (limbs
    drawn as colored ellipses, joints as colored dots), and a subset
    of the 478 face landmarks --- aligned with the dlib 68-point
    convention and extended with the FaceMesh lip contour --- is
    overlaid as white dots. This conversion bridges the format gap
    between the competition constraints and the ControlNet's
    pretrained input distribution, allowing the model to naturally
    respect the provided facial geometry. The control image is
    applied with conditioning strength 0.85.
    \item \textbf{PuLID}~\cite{PuLID2024}: injects target identity features
    with weight 0.85 to maximize identity similarity to the reference
    set. For each identity we construct the identity embedding by
    averaging PuLID encodings of up to 20 reference face crops selected
    from the available per-identity image pool via farthest-first
    diversity sampling on InsightFace embeddings (starting from the
    crop with the highest face-detection score and greedily adding the
    candidate that maximizes its minimum cosine distance to the
    already-selected set). The resulting embedding is cached once per
    identity and reused across all generations.
\end{enumerate}
Each generated image is downsampled from 512$\times$512 to 256$\times$256
using \texttt{cv2.INTER\_AREA}.

\begin{figure}[t]
\centering
\begin{subfigure}[t]{0.24\textwidth}
    \centering
    \includegraphics[width=\linewidth]{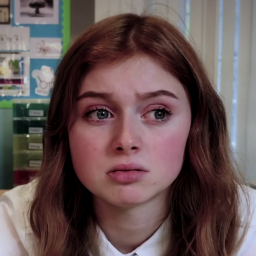}
    \caption{Reference crop\\(one of up to 20 per identity)}
    \label{fig:base-pipeline-a}
\end{subfigure}\hfill
\begin{subfigure}[t]{0.24\textwidth}
    \centering
    \includegraphics[width=\linewidth]{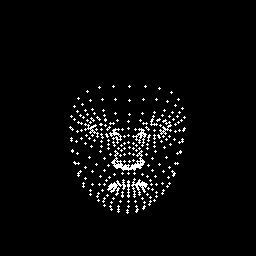}
    \caption{Competition landmark map (MediaPipe)}
    \label{fig:base-pipeline-b}
\end{subfigure}\hfill
\begin{subfigure}[t]{0.24\textwidth}
    \centering
    \includegraphics[width=\linewidth]{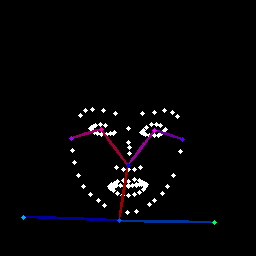}
    \caption{Our OpenPose-style control image}
    \label{fig:base-pipeline-c}
\end{subfigure}\hfill
\begin{subfigure}[t]{0.24\textwidth}
    \centering
    \includegraphics[width=\linewidth]{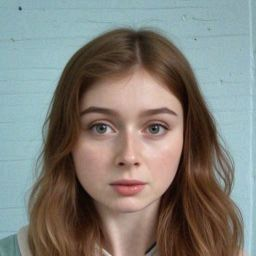}
    \caption{Final generated image}
    \label{fig:base-pipeline-d}
\end{subfigure}
\caption{Base generation pipeline for one identity (ID\_139, frame~3),
shown at the final $256{\times}256$ submission resolution.
(a) PuLID reference crop; (b) competition-supplied MediaPipe landmark
map; (c) our OpenPose-style ControlNet input; (d) final submission
image. Panel (d) was generated at $512{\times}512$ by FLUX.1-dev with
PuLID identity injection and OpenPose-style ControlNet conditioning,
downsampled to $256{\times}256$ via \texttt{cv2.INTER\_AREA}, and then
perturbed by our PGD adversarial attack
(Section~\ref{subsec:attack}); the $\epsilon = 2/255$ perturbation is
visually imperceptible. See Section~\ref{subsec:base-generation} for
component details.}
\label{fig:base-pipeline}
\label{fig:base-pipeline}
\end{figure}

\paragraph{Candidate selection.} For each identity we generate 11
candidates with different random seeds. We first remove candidates whose
landmark distance to the reference geometry is a clear outlier
(\texttt{lm\_dist} above a standard outlier threshold of 0.02, derived
from the per-identity score distribution); from the remaining candidates
we select the one with the highest identity similarity to the reference
anchor (Section~\ref{subsec:scoring}). After this automatic selection, a
small number of final images that still exhibited visible artifacts
(e.g., malformed or bent fingers) were manually replaced with the
next-best candidate by identity similarity among the remaining seeds.

\subsection{Adversarial Attack Design}
\label{subsec:attack}
The selected images are perturbed with a custom Projected Gradient
Descent (PGD)~\cite{PGD2018} attack jointly optimized against 12
white-box detectors: \textit{sdxl\_det}, \textit{effort},
\textit{haywood}, \textit{npr}, \textit{xception}, \textit{dfdc\_b7},
\textit{ateeq}, \textit{umm\_maybe}, \textit{f3net},
\textit{siglip\_dino}, \textit{freqnet}, and \textit{srm}. An additional
five detector groups (AIDE [3 checkpoints], UCF, RECCE, DistilDIRE,
SIDA-7B) are held out as black-box detectors used only for
transfer-evaluation (Section~\ref{subsec:gen-results}).

The attack minimizes a weighted sum of per-detector AI-class logits,
\begin{equation}
\mathcal{L}_{\text{adv}}(x) = \sum_{k=1}^{12} w_k \, f_k(x)_{\text{fake}},
\qquad \sum_k w_k = 1,
\end{equation}
where weights $w_k$ are tuned to balance per-detector difficulty. Updates
follow sign-based momentum PGD,
\begin{equation}
g_{t+1} = \mu \cdot g_t + \frac{\nabla_x \mathcal{L}_{\text{adv}}(x_t)}
{\lVert \nabla_x \mathcal{L}_{\text{adv}}(x_t) \rVert_1}, \qquad
x_{t+1} = \Pi_{\lVert x - x_0 \rVert_\infty \le \epsilon}
\!\left( x_t - \alpha \cdot \operatorname{sign}(g_{t+1}) \right),
\end{equation}
with $\epsilon = 2/255$, step size $\alpha = 0.5/255$, $T = 25$
iterations, and momentum decay $\mu = 1.0$.

Three additional components are inserted in the attack loop to improve
robustness and transferability:
\begin{itemize}
    \item \textbf{DiffJPEG-in-loop}~\cite{DiffJPEG2017}: a differentiable
    JPEG simulation at quality $Q = 80$ is applied to the perturbed
    image before each forward pass, so gradients reflect the
    post-compression image actually seen by detectors.
    \item \textbf{Input Diversity (DI)}~\cite{InputDiversity2019}: with
    probability $p = 0.5$ at each step the perturbed image is randomly
    resized to $r \in [224, 256]$ and zero-padded back to
    $256 \times 256$.
    \item \textbf{Expectation over Transformation
    (EoT)}~\cite{EoT2018}: realized implicitly by averaging
    gradients across the stochastic DI and DiffJPEG transformations.
\end{itemize}

\subsection{Challenges and Solutions}
\label{subsec:challenges}
\paragraph{Challenge 1: Modern ViT-based detectors resist conventional
post-processing.} Frequency-domain detectors can be evaded by JPEG
compression or histogram matching, but Vision Transformer detectors
built on CLIP~\cite{CLIP2021} or SigLIP~\cite{SigLIP2023} features proved robust
to such techniques. Our initial evasion rate on \textit{siglip\_dino}
was only 8\%. Switching to gradient-based PGD raised this to 96.9\%
(Table~\ref{tab:evasion}).

\paragraph{Challenge 2: Low-frequency attacks block SRM gradients.}
An early attempt applied Gaussian blur to the perturbation in order to
concentrate noise in low-frequency bands. This evaded frequency-based
detectors but completely zeroed out gradients through SRM-based
detectors (0\% evasion). Removing the blur and instead inserting
DiffJPEG~\cite{DiffJPEG2017} in the attack loop preserves gradients through
all detector types while keeping perturbations JPEG-robust. SRM evasion
recovered to 83.4\%.

\paragraph{Challenge 3: Transfer to unknown detectors.} White-box
attacks optimized for the 12 known detectors may not transfer to unseen
ones. We combined three transferability techniques in a single attack
loop --- Momentum Iterative (MI) gradient
accumulation~\cite{MomentumIterative2018}, Input Diversity
(DI)~\cite{InputDiversity2019}, and Expectation over Transformation
(EoT)~\cite{EoT2018} --- achieving 89--100\% evasion on the five
held-out black-box detector groups.

\paragraph{Challenges 4 \& 5: Per-detector failures and multi-detector
conflicts.} Individual detectors with stubborn failure cases were
addressed through (i) adaptive per-image weighting that doubles the loss
weight for images with $P(\text{fake}) > 0.5$, and (ii) two-stage
attacks that use the first-stage perturbation as a warm start for a
targeted second optimization. These recovered 104 images for
\textit{freqnet} and 31 images for \textit{f3net} / \textit{UCF}.

\paragraph{Challenge 6: Face-crop evasion is structurally infeasible.}
When detection pipelines crop the face region before analysis,
perturbations optimized over the full $256 \times 256$ image lose
effectiveness because spatial cropping breaks the pixel patterns the
perturbation depends on. Within the $\epsilon = 2/255$ budget we could
not simultaneously satisfy full-image and face-crop evasion. We did
observe, however, that the loss of effectiveness varied substantially
across detectors --- some retained their fooled prediction under
cropping while others recovered sharply --- and this asymmetry later
motivated the multi-detector response instability investigation of
Section~\ref{subsec:instability}. Robust solutions to the fundamental
$\ell_\infty$-bound limitation under spatial transformation are left
to future work.

\subsection{Internal Evaluation}
\label{subsec:scoring}
For both official image metrics we run the official competition
scoring code without modification; the only piece we supply ourselves
is the ground-truth identity embedding file, which the organizers do
not release. In addition, we evaluate detection-evasion performance
against the 17 detectors that we used inside the attack loop
(Section~\ref{subsec:attack}) and as held-out transfer verification,
summarised in Table~\ref{tab:evasion}.

\paragraph{Identity similarity.} The official scoring routine
(\texttt{process\_frames\_similarity}) extracts a single 512-d face
embedding from each submitted image using InsightFace
\texttt{buffalo\_s}~\cite{InsightFace2019}, L2-normalizes it, and
computes the cosine similarity against a per-identity ground-truth
embedding stored in \texttt{all\_embeddings\_images.json}. Because the
organizers do not release this ground-truth file, we construct it
ourselves: for each of the 335 identities, we extract face embeddings
with the same \texttt{buffalo\_s} model from multiple frames of the
corresponding development video, L2-normalize them individually, and
take the mean, yielding one 512-d unit-norm anchor per identity. When
the scoring routine cannot read an image or InsightFace fails to
detect a face in a submitted image, the per-image similarity defaults
to $-1$; the final score is the mean cosine similarity over all
$1{,}002$ submissions (including these $-1$ penalties). Under this
procedure with our self-constructed anchors, our submission attains a
mean identity distance ($1 - \text{mean cosine similarity}$) of
\textbf{0.2858}.

\paragraph{Landmark distance.} The official scoring routine
(\texttt{compare\_landmarks}) extracts 478 facial landmarks from each
submitted image with MediaPipe FaceMesh
(\texttt{refine\_landmarks=True}) and computes the mean per-landmark
Euclidean distance, in normalized image coordinates $[0,1]^2$, against
the target landmarks supplied by the organizers in the constraint
JSON. A default penalty of $0.5$ is substituted when the submitted
image is missing, when MediaPipe fails to detect a face in it, or
when the extracted and target landmark sets have mismatched shapes.
The final score is the mean of these per-image distances across all
$1{,}002$ submissions. Under this procedure, our submission attains a
mean landmark distance of \textbf{0.0065} (approximately 1.66 pixels
at the $256{\times}256$ submission resolution).

\paragraph{Detector evasion.}
\begin{table}[t]
\centering
\caption{Detector evasion rates for the 1{,}002 generated images. The
upper block lists the 12 white-box detectors used inside the attack
loop; the lower block lists the 5 held-out detector groups used only
for transfer evaluation.}
\label{tab:evasion}
\begin{tabular}{lrr}
\toprule
Detector & Evaded & Rate \\
\midrule
\multicolumn{3}{l}{\textit{White-box (attack targets)}} \\
sdxl\_det / effort / haywood / npr & 1002/1002 & 100\% \\
xception / dfdc\_b7 & 1001/1002 & 99.9\% \\
ateeq & 1000/1002 & 99.8\% \\
umm\_maybe & 981/1002 & 97.9\% \\
f3net & 975/1002 & 97.3\% \\
siglip\_dino & 971/1002 & 96.9\% \\
freqnet & 962/1002 & 96.0\% \\
srm & 836/1002 & 83.4\% \\
\midrule
\multicolumn{3}{l}{\textit{Transfer-verification (held-out black-box)}} \\
AIDE (3 checkpoints) & 1002/1002 & 100\% \\
UCF & 1002/1002 & 100\% \\
RECCE & 1000/1002 & 99.8\% \\
DistilDIRE & 953/1002 & 95.1\% \\
SIDA-7B & 893/1002 & 89.1\% \\
\bottomrule
\end{tabular}
\end{table}
Table~\ref{tab:evasion} summarizes evasion rates across all 17
detectors (12 white-box plus 5 held-out groups) that we used as
internal validation of our attack pipeline. We achieved $\geq 95\%$
evasion on 15 of 17 detectors; only \textit{srm} ($83.4\%$) and
\textit{SIDA-7B} ($89.1\%$) fell below this threshold.

\subsection{Score on Official Evaluation}
\label{subsec:gen-results}
The official evaluation scores across all sub-metrics are consolidated in
Table~\ref{tab:official-results} (Section~\ref{sec:official-results}).
In the official evaluation, our submission achieved an evasion score
of $\mathbf{0.9019}$ against organizer detectors ($90\%$ evaded) and
$\mathbf{0.5762}$ against participant detectors ($57.6\%$ evaded). The
gap between organizer ($90\%$) and participant ($57.6\%$) evasion
reflects the model-specific nature of adversarial attacks:
perturbations optimized against our 12 white-box targets transferred
well to organizer detectors (likely sharing architectural families
with our targets) but less effectively to participant detectors with
unknown architectures.

On the official image-quality side, the raw landmark distance was
$0.0065$ (identical to our internal measurement, since both rely on
the publicly supplied constraint landmarks) and the raw identity
distance was $0.5366$ (versus our internal $0.2858$ in
Section~\ref{subsec:scoring}; the gap reflects the difference between
our self-constructed reference anchors --- drawn from the same
development videos that supplied the constraint frames --- and the
organizers' held-out evaluation embeddings). After the official
scaling, these become a Landmark Distance Score of $\mathbf{0.7823}$
(concave normalization
$\max(0,\,1 - \sqrt{d / 0.137089})$ against a random-match baseline)
and an Identity Distance Score of $\mathbf{0.4553}$ (linear
normalization clipped between $0.4$ and $0.7$), averaged into an
Image Quality Score of $\mathbf{0.6188}$ and combined with the
evasion scores via the official rule
$\mathrm{Quality} \times (0.7\,\mathrm{Participants}
+ 0.3\,\mathrm{Organizers})$ to give a final task score of
$\mathbf{0.4170}$.

\section{Image Detection}
\label{sec:detection}

\subsection{Task Overview}
\label{subsec:det-task}
The detection task requires classifying $24{,}404$ images as Real
(label 0) or Fake (label 1), without any training set provided by
the organizers. Crucially, the Fake portion of the test set is itself
heterogeneous: it contains both \textbf{organizer-generated
face-manipulation deepfakes} (e.g., face-swap and reenactment outputs)
and \textbf{participant-generated AI-synthesized images} produced
under the parallel generation track. The official score is a weighted
combination of accuracy on organizer-generated deepfakes (weight
$0.1$), organizer-provided real data ($0.1$), curated real
ground-truth images ($0.4$), and participant-generated deepfakes
weighted by realism ($0.4$).

\subsection{Detection Framework}
\label{subsec:det-framework}
The two distinct fake types in the test set --- face manipulation and
full-image AI synthesis --- have different statistical signatures
(local face artifacts vs.\ global generation artifacts), and no
single off-the-shelf detector we surveyed covers both well. We
therefore deploy a \textbf{two-detector max-probability ensemble of
specialists}, one targeted at each fake type
(Figure~\ref{fig:detection-pipeline}):

\begin{figure}[t]
\centering
\includegraphics[width=\linewidth]{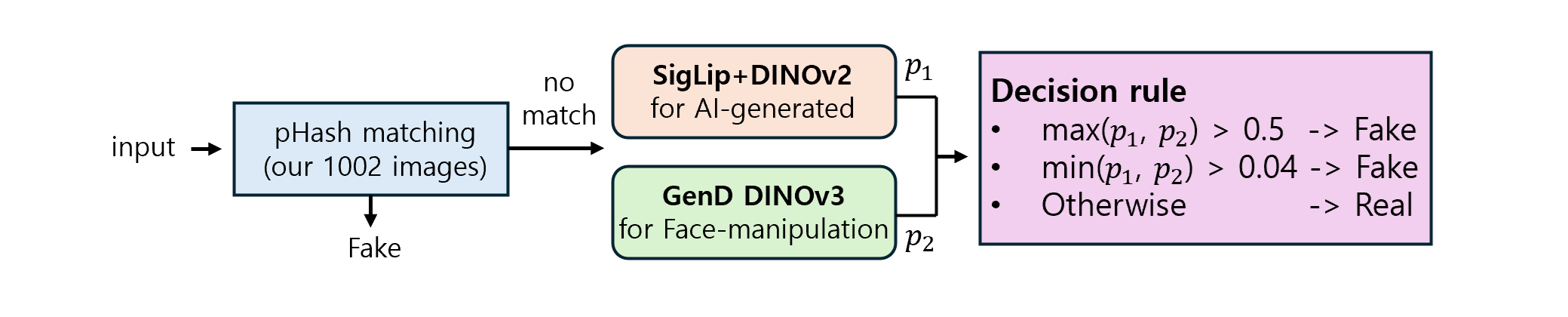}
\caption{Image detection pipeline. An input is first matched against
the perceptual hashes of our own $1{,}002$ generated submissions; an
exact match short-circuits to \textit{Fake}. Otherwise the image is
scored by two complementary detectors --- SigLIP+DINOv2 (AI-generated
content) and GenD-DINOv3 (face manipulation) --- and classified by the
decision rule of Section~\ref{subsec:threshold}.}
\label{fig:detection-pipeline}
\end{figure}

\begin{itemize}
    \item \textbf{SigLIP+DINOv2}~\mbox{\cite{SigLIP2023,BombekSigLIPDINOv2,DINOv2_2023}}:
    a publicly released HuggingFace checkpoint combining SigLIP-so400m
    and DINOv2 ViT-L dual encoders with LoRA~\cite{LoRA2022}
    fine-tuning, specialized for full-image AI-generated content
    detection.
    \item \textbf{GenD DINOv3}~\mbox{\cite{GenD2026,YermandyGenDDINOv3}}:
    a publicly released HuggingFace checkpoint with a DINOv3 ViT-L
    backbone and LayerNorm fine-tuning, specialized for face
    manipulation detection.
\end{itemize}

An image is first matched against a pre-computed perceptual hash
(pHash) set of our own $1{,}002$ generated submissions; an exact
match short-circuits to \textit{Fake}. Otherwise the image is scored
by both detectors and classified by a thresholded decision rule
combining their per-image probabilities $p_1$ (SigLIP+DINOv2) and
$p_2$ (GenD DINOv3): the image is \textit{Fake} if
$\max(p_1, p_2) > 0.5$, or if $\min(p_1, p_2) > \tau_{\min}$ with
$\tau_{\min} = 0.04$ (calibrated in
Section~\ref{subsec:threshold}); otherwise \textit{Real}.

\subsection{Model Selection}
\label{subsec:det-model-select}
Because no training set was provided, we compiled our own
$10{,}000$-image evaluation benchmark to compare candidate
pre-trained/fine-tuned detectors under conditions matching the
two expected fake types. The benchmark comprises $5{,}000$
face-cropped images from FaceForensics++~\cite{FF2019}
($1{,}000$ real frames plus five manipulation algorithms:
$1{,}000$ Deepfakes, $1{,}000$ Face2Face, $500$ FaceSwap,
$500$ FaceShifter, $1{,}000$ NeuralTextures) for the
face-manipulation regime, and $5{,}000$ images from BitMind
public pools\footnote{BitMind image pool of SDXL- and
FLUX-generated faces
(\url{https://huggingface.co/bitmind}), used here as a stand-in
for the unseen participant-generated AI imagery.}
($2{,}500$ SDXL and $2{,}500$ FLUX) for the AI-generation regime.

On this benchmark we evaluated four candidate detectors, each taken
off-the-shelf without further fine-tuning: ForAda~\cite{ForAda2025},
GenD CLIP~\mbox{\cite{GenD2026,YermandyGenDCLIP}},
GenD DINOv3~\mbox{\cite{GenD2026,YermandyGenDDINOv3}}, and the publicly
released LoRA fine-tuned SigLIP+DINOv2~\cite{BombekSigLIPDINOv2}
checkpoint. We measured AUROC separately on the two
face-manipulation sub-categories (\textit{Face Swap} =
FaceSwap+FaceShifter, \textit{Reenactment} =
Face2Face+NeuralTextures+Deepfakes) and the AI-generation set
(\textit{AI-gen} = BitMind).

\begin{table}[t]
\caption{Detection model comparison on our benchmark
($5{,}000$ FaceForensics++ + $5{,}000$ BitMind images).
DF=Deepfakes, F2F=Face2Face, FS=FaceSwap, FSh=FaceShifter,
NT=NeuralTextures; AI-gen aggregates $2{,}500$ BitMind SDXL and
$2{,}500$ BitMind FLUX. FM avg is the mean AUROC over the five
FF++ methods; Overall is the mean over all six columns.}
\label{tab:model_comparison}
\centering
\small
\setlength{\tabcolsep}{4pt}
\begin{tabular}{lcccccccc}
\toprule
\textbf{Model} & \textbf{DF} & \textbf{F2F} & \textbf{FS} & \textbf{FSh} & \textbf{NT} & \textbf{FM avg} & \textbf{AI-gen} & \textbf{Overall} \\
\midrule
ForAda~\cite{ForAda2025}                                       & 0.942 & 0.714 & 0.852 & 0.678 & 0.690 & 0.775 & 0.583 & 0.743 \\
GenD CLIP~\mbox{\cite{GenD2026,YermandyGenDCLIP}}              & \textbf{0.959} & 0.774 & \textbf{0.916} & 0.677 & 0.664 & 0.798 & 0.921 & \textbf{0.819} \\
\textbf{GenD DINOv3}~\mbox{\cite{GenD2026,YermandyGenDDINOv3}} & 0.930 & \textbf{0.810} & 0.877 & \textbf{0.678} & \textbf{0.754} & \textbf{0.810} & 0.724 & 0.796 \\
SigLIP+DINOv2~\cite{BombekSigLIPDINOv2}                        & 0.637 & 0.558 & 0.565 & 0.488 & 0.562 & 0.562 & \textbf{0.995} & 0.634 \\
\bottomrule
\end{tabular}
\end{table}

Table~\ref{tab:model_comparison} shows a clear specialization
pattern. GenD DINOv3 leads face-manipulation detection with the
highest FF++ average (FM avg $0.810$), while SigLIP+DINOv2
dominates AI-generation ($0.995$). GenD CLIP achieves the highest
Overall average ($0.819$) as a balanced generalist, but our
specialist ensemble surpasses it in both target regimes
(GenD DINOv3 $0.810 > 0.798$ on face manipulation; SigLIP
$0.995 > 0.921$ on AI-generation), justifying the two-detector
design of Section~\ref{subsec:det-framework}. Notably, FaceShifter
is universally the hardest FF++ method (AUROC $0.488$--$0.678$),
and FLUX-generated faces are harder to detect than SDXL for most
detectors, indicating that newer DiT-based generators are pushing
detection difficulty upward. We note that GenD DINOv3's absolute
AUROC on face manipulation shows some dependence on the face-crop
preprocessing pipeline, though it consistently ranks as the
strongest face-manipulation detector among the four models
evaluated.

\subsection{Threshold Calibration}
\label{subsec:threshold}
The default rule ``classify as Fake if either detector exceeds
$0.5$'' is conservative: an image can have both detectors weakly
flagging it (e.g., $p_1 = 0.3$, $p_2 = 0.2$) without being marked
Fake, even though their joint signal is suspicious. We therefore
introduced an auxiliary \texttt{prob\_min} rule --- classify as Fake
when $\min(p_1, p_2) > \tau_{\min}$ --- and calibrated $\tau_{\min}$
on a held-out $7{,}000$-image calibration set composed as follows:

\begin{itemize}
    \item \textbf{Real} ($2{,}000$): $1{,}500$ FF++ original frames
    + $500$ frames extracted from the generation-task reference
    videos.
    \item \textbf{Face manipulation, Fake} ($4{,}000$): FF++ five
    manipulation algorithms.
    \item \textbf{AI-generated, Fake} ($1{,}000$): $500$ images
    sampled from our own generation pipeline + $500$ BitMind
    SDXL/FLUX images.
\end{itemize}

Sweeping $\tau_{\min}$ on this set, we recorded the additional true
positives (TP) and false positives (FP) introduced by the
\texttt{prob\_min} rule on top of the $0.5$-baseline classification
(Table~\ref{tab:threshold_sweep}).

\begin{table}[t]
\caption{Threshold sweep for the \texttt{prob\_min} rule on the
calibration set ($2{,}000$ Real / $5{,}000$ Fake). ``+TP'' and
``+FP'' count the additional true and false positives produced by
the rule beyond the $0.5$-baseline. We selected
$\tau_{\min} = 0.04$ based on maximum net gain.}
\label{tab:threshold_sweep}
\centering
\small
\begin{tabular}{cccc}
\toprule
$\tau_{\min}$ & +TP & +FP & Net gain (TP $-$ FP) \\
\midrule
\textbf{0.04} & \textbf{825} & \textbf{392} & \textbf{433} \\
0.06 & 578 & 173 & 405 \\
0.10 & 356 &  94 & 262 \\
0.30 &  74 &  13 &  61 \\
\bottomrule
\end{tabular}
\end{table}

We selected $\tau_{\min} = 0.04$, which yields the largest count-based
net gain ($+433$) and raises overall calibration-set accuracy from
$78.0\%$ to $84.6\%$.

\subsection{Score on Official Evaluation}
\label{subsec:det-results}
The official evaluation scores across all sub-metrics are consolidated in
Table~\ref{tab:official-results} (Section~\ref{sec:official-results}).
In the official evaluation, our two-detector ensemble achieved
$99.4\%$ accuracy on organizer baseline deepfakes, $88.2\%$ on
participant-generated deepfakes, and $43$--$57\%$ on real images
across the curated and organizer-provided real subsets. Combined
under the official weighting
$0.1 \times \mathrm{OrgFake} + 0.1 \times \mathrm{OrgReal}
 + 0.4 \times \mathrm{CuratedReal}
 + 0.4 \times \mathrm{ParticipantFake}_{\text{realism-weighted}}$,
this yields a final image detection score of $\mathbf{0.6986}$.
Because real images carry the dominant $0.4$ weight, the weak
real-side accuracy substantially depressed the final score.

\paragraph{Post-competition analysis.}
Two compounding errors in our $\tau_{\min}$ calibration strategy
explain the real-image underperformance:

\begin{enumerate}
    \item \textbf{Inappropriate metric.} We optimized count-based net
    gain (TP $-$ FP), which implicitly assumes the test set's class
    proportions match the calibration set's. Expressed as rates on
    the calibration set, however, the $\tau_{\min}=0.04$ rule produces
    a true-positive-rate gain of $825/5000 = 16.5\%$ while incurring
    a false-positive rate of $392/2000 = 19.6\%$ --- a higher rate of
    new false positives than new true positives. The rule looked
    profitable in raw counts only because Fake outnumbered Real
    $5{,}000 : 2{,}000$ in our calibration set.
    \item \textbf{Distribution mismatch.} Our calibration set had a
    Real$:$Fake ratio of $1{:}2.5$, but the official test set, given
    the $0.4$ weight on curated real images, contains a higher real
    fraction. Because the rule's $19.6\%$ FPR exceeds its $16.5\%$
    TPR gain (item 1), at any balanced or real-dominant test
    distribution the new false positives outweigh the new true
    positives, turning the rule into a net loss regardless of
    absolute test-set size.
\end{enumerate}

The principled remedy would have been to calibrate per-detector
thresholds against the ROC curve of each detector on the FF++
benchmark, or to use rate-independent target metrics
(TPR / FPR / balanced accuracy) instead of count-based net gain.
Given that our AI-generation specialist SigLIP+DINOv2 achieves
AUROC $0.995$ on BitMind and our face-manipulation specialist
GenD DINOv3 averages $0.810$ across all five FaceForensics++
methods (Table~\ref{tab:model_comparison}), correct threshold
selection alone is expected to lift overall accuracy above $90\%$.

\section{Official Competition Results}
\label{sec:official-results}

Table~\ref{tab:official-results} consolidates our official CLEF 2026
ImageCLEF Deepfake Task scores across both sub-tasks; per-task
derivations and post-mortem discussion appear in
Sections~\ref{subsec:gen-results} and~\ref{subsec:det-results}.

\begin{table}[t]
\centering
\caption{Official CLEF 2026 ImageCLEF Deepfake Task results for team
``Go To Germany''.
\textbf{Generation final score} $= \mathrm{Quality} \times
(0.7\,\mathrm{Participants} + 0.3\,\mathrm{Organizers})$, where
Quality is the mean of the Landmark and Identity distance scores.
\textbf{Detection final score} is a weighted sum with weights
$0.1$, $0.1$, $0.4$, $0.4$ over Baseline Deepfakes,
Organizers Real, Organizers Real GT, and realism-weighted
Participant Deepfakes, respectively. The high-weight components
(participant evasion, curated real) are the primary drivers of the
final scores.}
\label{tab:official-results}
\small
\begin{tabular}{lr}
\toprule
\textbf{Metric} & \textbf{Score} \\
\midrule
\multicolumn{2}{l}{\textit{Generation Task} (Section~\ref{sec:generation})} \\
\quad Landmark distance score          & 0.7823 \\
\quad Identity distance score          & 0.4553 \\
\quad Image Quality Score (mean)       & 0.6188 \\
\quad Evasion, Organizer detectors     & 0.9019 \\
\quad Evasion, Participant detectors   & 0.5762 \\
\quad \textbf{Final Generation Score}  & \textbf{0.4170} \\
\midrule
\multicolumn{2}{l}{\textit{Detection Task} (Section~\ref{sec:detection})} \\
\quad Baseline Deepfakes                                 & 0.9939 \\
\quad Organizers Real Data                               & 0.4324 \\
\quad Organizers Real GT Data (curated)                  & 0.5729 \\
\quad Participant Deepfake Data (unweighted, info.)      & 0.8822 \\
\quad Participant Deepfake Data, realism-weighted        & 0.8171 \\
\quad \textbf{Final Detection Score}                     & \textbf{0.6986} \\
\bottomrule
\end{tabular}
\end{table}

The benchmark's scoring rules concentrate weight on components that
require generalization beyond the organizer-provided baselines: on
the generation side, participant-side evasion carries $0.7$ weight
within the evasion factor (versus $0.3$ for organizer-side evasion);
on the detection side, a held-out curated real subset carries the
single largest weight ($0.4$). Our submission scores highly on the
lower-weight components of both tasks --- $0.9019$ organizer
evasion, $0.9939$ accuracy on baseline organizer deepfakes --- but
is bounded by the high-weight components ($0.5762$ participant
evasion, $0.5729$ curated-real accuracy). The final scores of
$0.4170$ (generation) and $0.6986$ (detection) therefore reflect the
difficulty of the benchmark's explicit generalization objective
rather than raw capability limits on the organizer-only regime.

Reading across the two tasks, a common thread connects them to our
further experiments. On the generation side, we demonstrably drove
the fake-class probability of most off-the-shelf detectors below
the decision threshold using standard adversarial post-processing
techniques. On the detection side, our ensemble's accuracy dropped
from $0.9939$ on baseline organizer deepfakes to $0.8822$ on
participant-generated deepfakes --- a gap most plausibly explained
by distribution shift between the two generation pipelines, with
adversarial post-processing as a possible secondary contributor
if participants applied comparable techniques. Since our own
generation track directly demonstrates the adversarial
vulnerability affecting raw-pixel detectors, Section~\ref{sec:purification}
investigates whether the same off-the-shelf detectors can be used
to \emph{flag} adversarial-post-processed inputs as such.

\section{Further Experiments: Purification-Based Adversarial Detection}
\label{sec:purification}

Building on the observation of Section~\ref{sec:official-results} that
adversarial post-processing bounds detector accuracy regardless of
representational backbone, we asked whether our own detectors can
\emph{flag} such inputs --- not necessarily recover the correct
fake-vs-real label, but distinguish them from natural, clean, or
naturally-degraded inputs.

We explored three families of detection signals in sequence:
\emph{multi-detector response instability}
(Section~\ref{subsec:instability}), \emph{feature-space Mahalanobis
outlier detection} (Section~\ref{subsec:maha}), and \emph{input-space
purification response} (Section~\ref{subsec:purification-input}). The
first two yielded informative failure modes; purification is where
our final positive result is concentrated.

\subsection{Multi-Detector Response Instability}
\label{subsec:instability}

\paragraph{Motivation from Challenge~6.}
Challenge~6 of Section~\ref{subsec:challenges} reported that
face-cropping an adversarial image before inference changed the
perturbation's effectiveness \emph{differently across detectors}.
This suggested that adversarial inputs leave a signature in the
\emph{joint} response of multiple detectors to controlled input
modifications, even when no single detector's modified output is
individually informative.

\paragraph{Method.}
For each input $x$ we generated three perturbed variants --- center
crop $192 \times 192$, random crop $192 \times 192$, and a face-aware
crop (variable size around the detected face) --- and passed each
through a panel of eight detectors: AIDE (three checkpoints: progan,
genimage, sd14), DistilDIRE, NPR, Haywood, Xception, and
SigLIP+DINOv2. The resulting $8 \times 3 = 24$-dimensional shift
vector was the input to logistic regression (LR) and random forest
(RF) classifiers trained on a $1{,}001$-image train /
$1{,}001$-image test split. Adversarial inputs are stored as PNG but
generated through the JPEG-$Q80$-in-loop PGD pipeline of
Section~\ref{subsec:attack}, so the perturbations are JPEG-survivable
by construction.

\paragraph{Initial results.}
Single-detector AUROCs spanned a wide range
(Table~\ref{tab:instability-auc}), with the best configuration at
$0.761$ and several sub-random (signal direction reversed). The
combined classifiers reached \textbf{LR AUROC $0.971$ / RF AUROC
$0.981$} (accuracies $0.914$ / $0.932$) with clean-fake false-positive
rate below $0.5\%$. A controlled discrimination test on $400$
adversarials vs.\ $400$ real images JPEG-compressed at
$Q\in\{50, 60, 70, 80\}$ reached RF AUROC $0.952$, further suggesting
the signal could be made JPEG-robust with augmented training.

\begin{table}[h]
\centering\small
\caption{Section~\ref{subsec:instability} --- per-detector AUROC of
single-variant probability shift (selected entries) and combined-vector
classifier AUROC on the clean-vs-adversarial split. Single detectors
give weak and sometimes reversed signal; combining the $24$ features
by LR or RF yields a strong adversarial detection signal.}
\label{tab:instability-auc}
\begin{tabular}{lcc}
\toprule
Configuration & AUROC & Accuracy \\
\midrule
aide\_genimage / center crop (best single) & 0.761 & --- \\
distildire / face crop                      & 0.761 & --- \\
aide\_genimage / random crop                & 0.757 & --- \\
npr / face crop                             & 0.663 & --- \\
distildire / center crop                    & 0.557 & --- \\
aide\_progan / random crop                  & 0.507 & --- \\
aide\_sd14 / random crop                    & 0.338 & --- \\
haywood / random crop                       & 0.338 & --- \\
\midrule
\textbf{Logistic regression (24-dim)}       & \textbf{0.971} & 0.914 \\
\textbf{Random forest (24-dim)}             & \textbf{0.981} & 0.932 \\
\bottomrule
\end{tabular}
\end{table}

\paragraph{Diagnosis: JPEG re-encoding mimics the adversarial signature.}
The pilot did not survive deployment conditions. A control experiment
applied nine augmentations --- Gaussian noise
($\sigma\in\{0.02, 0.05\}$), Gaussian blur ($k\in\{3, 5\}$), JPEG
re-encoding ($Q\in\{50, 70\}$), rotation ($\{15^\circ, 20^\circ\}$),
and Poisson noise --- to twenty real images and measured the cosine
similarity of each instability vector to the adversarial-suspect
cluster (Table~\ref{tab:instability-confound}). JPEG re-encoding at
any moderate quality produced cosine similarity above $0.92$,
indistinguishable from the adversarial signature; all other
augmentations stayed below $0.90$.

\begin{table}[h]
\centering\small
\caption{Section~\ref{subsec:instability} --- cosine similarity of
augmented-real instability vectors to the adversarial-suspect cluster.
JPEG re-encoding at $Q\in\{50, 70, 80\}$ produces signatures
indistinguishable from adversarial ($\cos > 0.92$), while natural
noise and blur augmentations stay below $0.70$.}
\label{tab:instability-confound}
\begin{tabular}{lc}
\toprule
Augmentation                       & $\cos$ to adversarial signature \\
\midrule
\textbf{JPEG Q70}                  & \textbf{0.964} \\
\textbf{JPEG Q80}                  & \textbf{0.952} \\
\textbf{JPEG Q50}                  & \textbf{0.921} \\
Gaussian noise ($\sigma{=}0.02$)   & 0.890 (grey) \\
Rotation $20^\circ$                & 0.826 (grey) \\
Rotation $15^\circ$                & 0.817 (grey) \\
Gaussian blur ($k{=}3$)            & 0.765 (grey) \\
Gaussian noise ($\sigma{=}0.05$)   & 0.682 (distinct) \\
Gaussian blur ($k{=}5$)            & 0.674 (distinct) \\
Poisson noise                      & 0.642 (distinct) \\
\bottomrule
\end{tabular}
\end{table}

A deployment check on the full $24{,}404$-image test set confirmed
the consequence. The organizers released the test images without
processing metadata, so we estimated per-image JPEG quality from the
block-artifact ratio (BAR). On this estimate, the random-forest
adversarial probability correlated negatively with BAR-estimated
quality (Spearman $\rho = -0.519$, $p < 10^{-50}$) and the
false-positive rate scaled monotonically with estimated compression
strength (Table~\ref{tab:instability-fp-by-quality}), from $87.2\%$
at estimated $Q<50$ down to $6.7\%$ at estimated $Q\geq 90$. Combined
with the controlled-augmentation result above
(Table~\ref{tab:instability-confound}), this indicates that the
instability signal is entangled with JPEG-like block artifacts and
cannot serve as an adversarial-specific decision on test inputs whose
compression history is unknown. This motivated the move to a
single-detector feature-space signal (Section~\ref{subsec:maha}) that
could distinguish adversarial perturbations from such structured
high-frequency content.

\begin{table}[h]
\centering\small
\caption{Section~\ref{subsec:instability} --- adversarial false-positive
rate on the $24{,}404$-image deployment test set, binned by estimated
JPEG quality (random forest, threshold $0.5$). Low-quality JPEG real
images trigger almost-certain false positives, revealing that the
instability signal is structurally entangled with JPEG quantization
artifacts.}
\label{tab:instability-fp-by-quality}
\begin{tabular}{lrr}
\toprule
JPEG quality bin & $n$ & False-positive rate \\
\midrule
$Q < 50$           & 556  & 87.2\% \\
$Q \in [50, 60)$   & 618  & 71.7\% \\
$Q \in [60, 70)$   & 469  & 43.1\% \\
$Q \in [70, 80)$   & 823  & 23.0\% \\
$Q \in [80, 90)$   & 1091 & 10.5\% \\
$Q \geq 90$        & 2445 & \phantom{0}6.7\% \\
\bottomrule
\end{tabular}
\end{table}

\subsection{Mahalanobis Feature-Space Outlier Detection}
\label{subsec:maha}

\paragraph{Hypothesis.}
The instability signal of Section~\ref{subsec:instability} failed
because the JPEG signature is empirically indistinguishable from the
adversarial signature at the level of multi-detector probability
shifts. We hypothesized that a \emph{single-detector} signal computed
on the detector's internal features might separate the two:
adversarial inputs are optimized against the classification head and
should land in low-likelihood regions of the per-class feature
distribution, while JPEG-compressed clean inputs --- though
high-frequency-distorted --- should still sit on the natural-image
feature manifold. Mahalanobis distance to per-class Gaussian fits, following the adversarial / out-of-distribution detection framework of Lee et al.~\cite{Mahalanobis2018}, is the natural test of this
off-manifold property.

\subsubsection{Per-detector single-distance Mahalanobis}
\label{subsubsec:maha-single}

Throughout Section~\ref{subsec:maha} we use four image categories:
\textbf{real} ($3{,}350$ face crops from the generation-task reference
videos), \textbf{clean fake} (our $1{,}002$ FLUX+PuLID generations
before PGD), \textbf{JPEG-Q70/Q80} (the same real frames re-encoded
at the indicated quality), and \textbf{adversarial} (our $1{,}002$
PGD-perturbed submission images).

For both SigLIP+DINOv2 and ForAda we fitted per-class Gaussians on
the penultimate features and scored each test image by the minimum
Mahalanobis distance to either class centroid,
$\min(d_{\text{real}}, d_{\text{fake}})$. The result repeated the
Section~\ref{subsec:instability} failure mode
(Table~\ref{tab:maha-single-auc}): adv-vs-all is barely above chance
(AUROC $0.515$ on SigLIP), and adv-vs-JPEG is near-zero on both
detectors and both quality levels --- a near-perfect \emph{reverse}
signal in which JPEG-compressed reals sit \emph{farther} from the
real-class centroid than adversarial images do.

\begin{table}[h]
\centering\small
\caption{Section~\ref{subsubsec:maha-single} --- AUROC of
$\min d_{\text{maha}}$ as a single-distance adversarial detector on
two detectors. On both adv-vs-JPEG comparisons the score is
near-zero, meaning JPEG-compressed real images are pushed
\emph{farther} into the off-manifold region than adversarial images
are. The same JPEG confound that defeated multi-detector instability
re-emerges at the feature level.}
\label{tab:maha-single-auc}
\begin{tabular}{lccc}
\toprule
Detector & adv vs all & adv vs JPEG-Q60 & adv vs JPEG-Q80 \\
\midrule
SigLIP+DINOv2 & 0.515 & 0.022 & 0.182 \\
ForAda        & 0.427 & 0.010 & 0.072 \\
\bottomrule
\end{tabular}
\end{table}

JPEG quantization predominantly affects shallow-layer features (where
high-frequency content lives), while the adversarial signal is
structured to fool the classification head and should concentrate
near deep layers. A multi-layer pooled distance should therefore
amplify the adversarial-specific component and suppress the
JPEG-driven shallow component.

\subsubsection{Multi-layer Mahalanobis ensemble}
\label{subsubsec:maha-multilayer}

Following Lee et al.~\cite{Mahalanobis2018}, we extracted DINOv2
features at five depths ($L_4, L_9, L_{14}, L_{19}, L_{23}$), fitted
per-class Gaussians at each depth with Ledoit-Wolf shrinkage
covariance, and combined the five per-layer Mahalanobis distances
via logistic regression on a balanced pilot set of $100$ images per
category (real, clean fake, JPEG-Q80, adversarial; sources as
defined in Section~\ref{subsubsec:maha-single}).

\begin{figure}[t]
\centering
\begin{subfigure}[t]{0.85\textwidth}
    \centering
    \includegraphics[width=\linewidth]{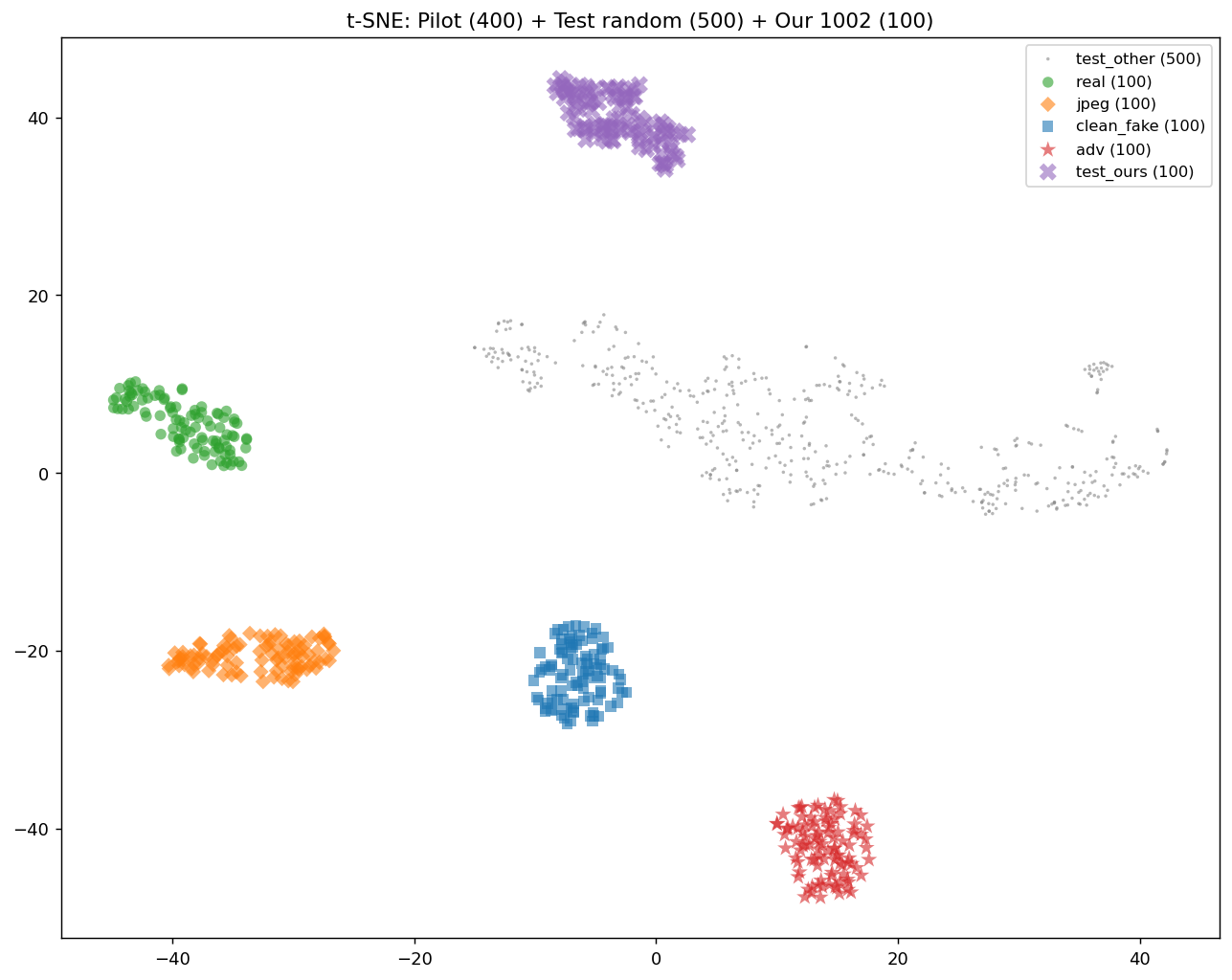}
    \caption{Pilot t-SNE: $400$ pilot points (100 each from the four
    categories defined in Section~\ref{subsubsec:maha-single}) plus
    $500$ random points from the ImageCLEF test set and our $100$
    submitted adversarials. Markers: clean fakes (blue squares),
    JPEG-Q80 reals (orange diamonds), adversarials (red stars).}
    \label{fig:maha-tsne-pilot}
\end{subfigure}

\vspace{0.5em}

\begin{subfigure}[t]{0.85\textwidth}
    \centering
    \includegraphics[width=\linewidth]{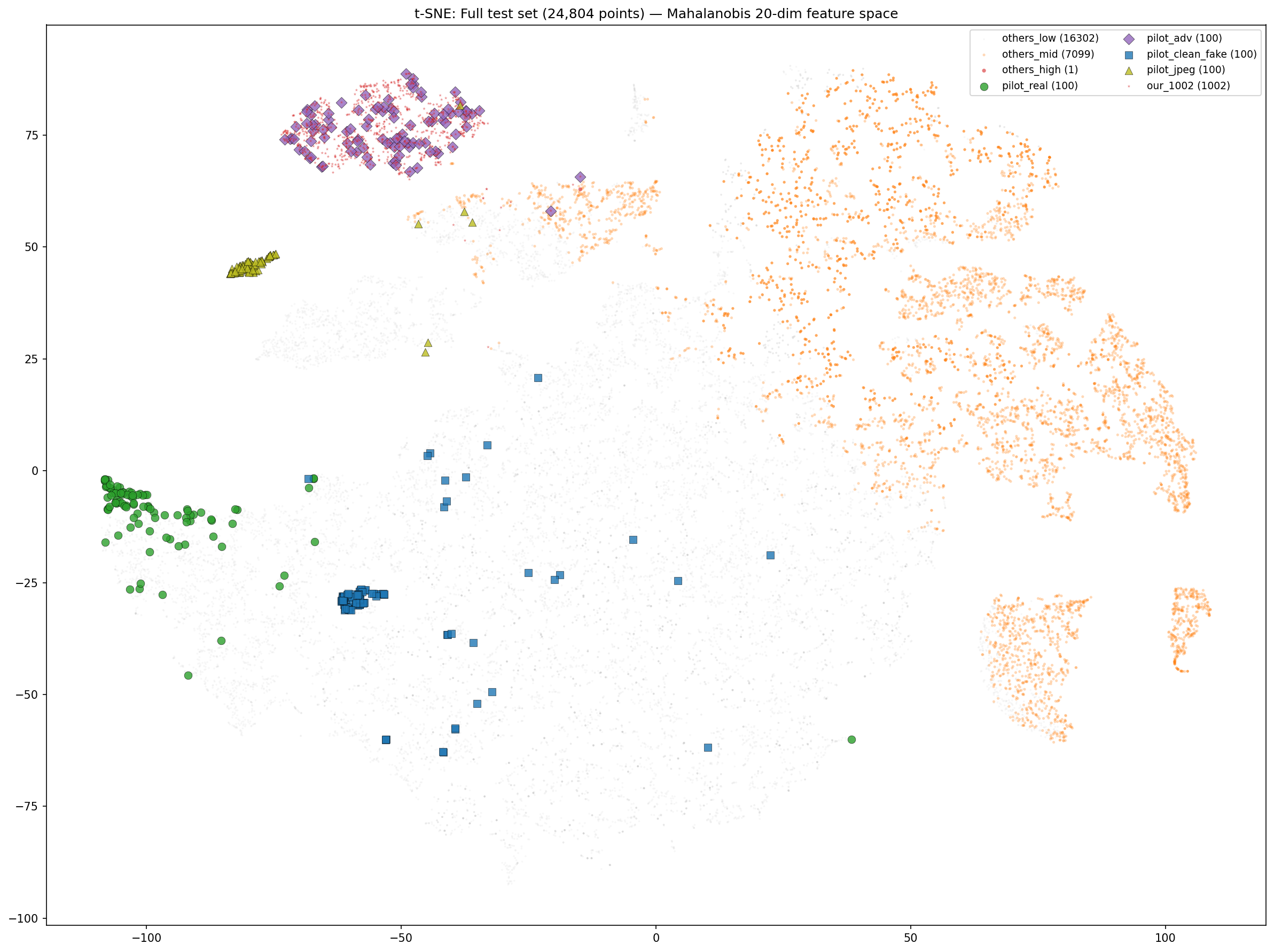}
    \caption{Full $24{,}404$-image ImageCLEF test set overlaid on
    the same Mahalanobis $20$-dimensional feature space, with the
    pilot category clusters retained as small markers. Test points
    are binned by the LR ensemble's predicted adversarial
    probability $p$: \textit{others\_low} ($p < 0.4$, grey dots),
    \textit{others\_mid} ($0.4 \leq p \leq 0.7$, orange circles),
    and \textit{others\_high} ($p > 0.7$, red squares).}
    \label{fig:maha-tsne-test}
\end{subfigure}

\caption{Section~\ref{subsubsec:maha-multilayer} --- multi-layer
Mahalanobis t-SNE on the pilot (a) and after refitting on the full
test feature corpus (b).}
\label{fig:maha-tsne}
\end{figure}

\paragraph{Pilot.}
The LR ensemble reached \textbf{AUROC $1.000$} on all three reference
sets (real, clean fake, JPEG) --- the strongest single-signal result
of our exploration (Table~\ref{tab:maha-multilayer-auc}). Single-layer
failures at $L_4$ and $L_{14}$ on adv-vs-JPEG were absorbed by the
ensemble. A pilot t-SNE projection (Figure~\ref{fig:maha-tsne}a)
shows four cleanly separated clusters --- the first time in this
investigation that adversarial and JPEG-Q80 inputs separated in a
learned feature space.

\begin{table}[h]
\centering\small
\caption{Section~\ref{subsubsec:maha-multilayer} --- per-layer DINOv2
Mahalanobis AUROC on the pilot (100 images per category). Single
layers fail catastrophically at $L_4$ and $L_{14}$ on adv-vs-JPEG,
but the LR ensemble of all five depths reaches AUROC $1.0$ across
all three reference sets --- the strongest pilot result of our
exploration.}
\label{tab:maha-multilayer-auc}
\begin{tabular}{lccc}
\toprule
Layer & real vs adv & adv vs JPEG & adv vs clean \\
\midrule
$L_4$                & 1.000 & 0.463 & 1.000 \\
$L_9$                & 1.000 & 1.000 & 1.000 \\
$L_{14}$             & 1.000 & 0.193 & 1.000 \\
$L_{19}$             & 1.000 & 1.000 & 1.000 \\
$L_{23}$             & 1.000 & 0.981 & 0.923 \\
\textbf{LR ensemble} & \textbf{1.000} & \textbf{1.000} & \textbf{1.000} \\
\bottomrule
\end{tabular}
\end{table}

\paragraph{Deployment failure.}
The pilot did not survive the move to the full $24{,}404$-image test
set. We refit the per-class Gaussians on the full pilot+test feature
corpus and re-evaluated the LR ensemble. CV AUROC remained $1.000$,
but \textbf{test-set detection rate dropped to $0.0\%$} and
\textbf{recall on our own $1{,}002$ submitted adversarials was also
$0.0\%$}. The full-data Gaussian has substantially broader covariance
than the pilot Gaussian: mean Mahalanobis distance is
$\mathbf{6.75\times}$ larger ($1{,}514$ vs.\ pilot $224$), so the LR
boundary trained on pilot-scale distances is meaningless at the
deployment scale. Five rescue attempts --- Gaussian refit on test
features, per-dimension z-score normalization, raw-distance
thresholding, inter-class distance ratios, and inverted-distance
classification --- all yielded $\leq 4.4\%$ recall on our own
adversarials. The pilot t-SNE separation
(Figure~\ref{fig:maha-tsne}a) is qualitatively gone at deployment
scale (Figure~\ref{fig:maha-tsne}b).

\paragraph{Diagnosis.}
The multi-layer ensemble \emph{did} resolve the JPEG confound of
Sections~\ref{subsec:instability} and~\ref{subsubsec:maha-single} at
pilot scale --- the first signal in our investigation to do so. But
any Mahalanobis-based score requires a fixed-reference Gaussian whose
statistics depend on the calibration data, and when the deployment
distribution is broader than the calibration distribution the
distance scale shifts and any pilot-calibrated decision boundary
loses meaning. We therefore abandoned feature-space outlier signals
in favor of an input-space approach
(Section~\ref{subsec:purification-input}) that measures detector
\emph{behaviour} under a controlled input change rather than fitting
a statistical model over features --- a formulation that does not
depend on calibration-scale matching.

\subsection{Input-Space Purification}
\label{subsec:purification-input}

The Mahalanobis ensemble of Section~\ref{subsec:maha} collapsed at
deployment because its decision boundary was tied to calibration-scale
Gaussian statistics. We therefore moved to signals that require no
fitted reference: \emph{input-space purification}. The hypothesis is
that detectors respond differently to controlled image filters on
adversarially perturbed inputs than on natural inputs, because
adversarial perturbations are gradient-aligned and lose their effect
once the gradient-relevant pixel pattern is smoothed away. We
investigated this in three steps of increasing scope: a single-operator
pilot on the two-detector ensemble (Section~\ref{subsubsec:pur-single}),
a multi-operator combination on a single detector
(Section~\ref{subsubsec:pur-multi}), and a fine-tuning ablation across
six detectors sharing the same backbone
(Section~\ref{subsubsec:pur-ablation}).

\subsubsection{Single-operator median-3 on the two-detector ensemble}
\label{subsubsec:pur-single}

The simplest instantiation applied a $3 \times 3$ median filter to
each input and recorded the change in detector probability,
$\Delta = \sigma^{-1}(p(T(x))) - \sigma^{-1}(p(x))$, where
$\sigma^{-1}$ is the logit function. We evaluated this score on the
two-detector ensemble of Section~\ref{subsec:det-framework}
(SigLIP+DINOv2 and GenD-DINOv3) across five categories of $100$
images each: clean real, clean fake, JPEG-$Q80$ compressed reals,
adversarial against ForAda, and adversarial against SigLIP+DINOv2.

\begin{table}[h]
\centering\small
\caption{Section~\ref{subsubsec:pur-single} --- single-operator
median-$3$ purification AUROC on the two-detector ensemble. Each
own-target adversarial separates well from clean fakes on the target
detector itself, but the signal collapses on other detectors and on
the JPEG-$Q80$ contrast.}
\label{tab:pur-single}
\begin{tabular}{lcccc}
\toprule
& adv\_forada & adv\_forada & adv\_siglip & adv\_siglip \\
Detector & vs clean & vs JPEG & vs clean & vs JPEG \\
\midrule
SigLIP+DINOv2 & 0.872 & 0.225 & \textbf{0.997} & 0.834 \\
GenD\_DINOv3  & 0.481 & 0.496 & 0.384 & 0.385 \\
ForAda        & 0.642 & 0.554 & 0.241 & 0.183 \\
EFFORT        & 0.093 & 0.693 & 0.088 & 0.701 \\
\midrule
Ensemble adv-vs-JPEG (combined) & \multicolumn{4}{c}{0.530} \\
\bottomrule
\end{tabular}
\end{table}

SigLIP+DINOv2 separates its own-target adversarials cleanly from
clean fakes (AUROC $0.997$), but JPEG-$Q80$-compressed reals dilute
the ensemble signal back to chance ($0.530$) --- the same JPEG
confound that defeated Sections~\ref{subsec:instability}
and~\ref{subsec:maha}, now reappearing in the input domain. This
motivates combining purification operators whose responses to
adversarial perturbation and JPEG quantization differ.

\subsubsection{Multi-operator combination on GenD-DINOv3}
\label{subsubsec:pur-multi}

We applied three input filters (median $5{\times}5$, bit-depth $4$,
Gaussian $\sigma{=}1$) and recorded the feature shift each induced
at layer $L_{19}$ of GenD-DINOv3, plus the raw output $\Delta$-logit
as a fourth feature. A logistic regression classifier on this
four-dimensional response vector reached AUROC
$\mathbf{0.998 \pm 0.004}$ on adversarial vs.\ JPEG-$Q80$ (5-fold CV)
and $0.932 \pm 0.064$ on adversarial vs.\ all non-adversarial inputs.

\begin{table}[h]
\centering\small
\caption{Section~\ref{subsubsec:pur-multi} --- logistic regression
coefficients on the four-dimensional operator response vector at
GenD-DINOv3's $L_{19}$. Median and bit-depth coefficients are
negative; Gaussian and raw-$\Delta$logit are positive.}
\label{tab:pur-multi-coefs}
\begin{tabular}{lr}
\toprule
Feature & Coefficient \\
\midrule
$\Delta_{L_{19}}$ (median $5\times5$)    & $-2.01$ \\
$\Delta_{L_{19}}$ (bit-depth 4-bit)      & $-1.96$ \\
$\Delta_{L_{19}}$ (Gaussian $\sigma{=}1$) & $+1.41$ \\
$\Delta$-logit (raw)                      & $+1.21$ \\
\bottomrule
\end{tabular}
\end{table}

The sign pattern in Table~\ref{tab:pur-multi-coefs} is what the
ensemble exploits: median and bit-depth reduce adversarial
high-frequency content but leave JPEG quantization residuals
relatively untouched, while the Gaussian operator and the raw output
shift in the same direction on both input types. The signed
combination amplifies the difference between the two regimes.

This resolves the JPEG confound on GenD-DINOv3, but the LR weights
must be refit per detector and per deployment distribution. The
next section asks whether a simpler, weight-free signal can yield a
JPEG-robust adversarial detector across an ablation of fine-tuning
methods.

\subsubsection{Raw $|\Delta\text{logit}|$ with fine-tuning ablation}
\label{subsubsec:pur-ablation}

\paragraph{Hypothesis.}
A simpler, weight-free signal --- raw $|\Delta\text{logit}|$ from a
single purification operator --- would remove the per-detector and
per-distribution recalibration constraints of
Section~\ref{subsubsec:pur-multi}, provided the underlying detector
is sensitive enough to the adversarial perturbation. We hypothesized
that this sensitivity is controlled by how strongly the detector's
fine-tuning modifies the pretrained backbone: a backbone-preserving
detector should retain the smooth input-output mapping that makes
gradient-aligned perturbations stand out under purification, while a
heavily fine-tuned backbone may have smoothed over that mapping. We
tested this with a controlled ablation across six detectors sharing
the same CLIP ViT-L/14 backbone.

\paragraph{Setup.}
We trained three detectors ourselves on an identical FaceForensics++
split with all other hyperparameters matched, differing only in
fine-tuning method: a linear probe (LP, $0\%$ of backbone parameters
updated), a LoRA fine-tune ($r{=}32$, $\approx 1\%$ updated), and a
full fine-tune ($\approx 100\%$). To test cross-recipe
generalization, we added three publicly released checkpoints sharing
the same backbone but with different fine-tuning strategies and
training data: GenD-CLIP (LayerNorm-only), ForAda (adapter modules),
and EFFORT~\cite{EFFORT2025} (SVD-residual decomposition). The detection signal is the
raw output logit shift under median-$3$ purification,
$|\,f(T(x)) - f(x)\,|$, computed in the pre-sigmoid logit domain. We
read this from the final fake-class output rather than an
intermediate feature because the six detectors share a backbone but
differ in fine-tuning, so no intermediate layer is comparable across
them.

\paragraph{Pilot and metric correction.}
On a balanced $100$-image-per-category pilot the backbone-preserving
methods showed strong separation: the linear probe shifted by mean
$+31.06$ logits on adversarial inputs versus $+0.85$ on real inputs,
a $36\times$ ratio. This signal is invisible in the sigmoid-saturated
metric typically reported in prior work --- after $\sigma(\cdot)$
followed by $\sigma^{-1}(\cdot)$ with floating-point clipping, the
$+31$ raw shift compresses to $+0.30$, smaller than the $+0.85$
apparent shift on real inputs. The discrepancy is purely
metric-induced; we report raw logit shifts throughout the
validations below.

\paragraph{Deployment validation.}
We tested the pilot signal on a deployment matrix of four
adversarial source types $\times$ two input degradations. The source
types are: \emph{production} (our $1{,}002$ submitted images, PGD
against a 12-detector ensemble with DiffJPEG-in-loop at $Q{=}80$);
\emph{cross AI-generated} (BitMind SDXL/FLUX and our generation
pipeline, attacked with PGD~\cite{PGD2018}/FGSM~\cite{FGSM2015}/BIM~\cite{BIM2017}
targeting forada/SigLIP/joint ensembles); \emph{cross face-swap} (FF++ Deepfakes, FaceShifter,
FaceSwap with the same attack matrix); and \emph{cross reenactment}
(FF++ Face2Face, NeuralTextures with the same attack matrix). The
two degradations are raw PNG input and post-JPEG-$Q80$ input. AUROC
is reported with bootstrap $95\%$ confidence intervals over $5{,}000$
resamples (Table~\ref{tab:pur-ablation-main}).

\begin{table}[h]
\centering\small
\caption{Section~\ref{subsubsec:pur-ablation} --- raw
$|\Delta\text{logit}|$ AUROC for adversarial detection across six
detectors, four source types, and two input degradations. EFFORT
generalizes broadly under raw input (AUROC $0.81$--$0.98$ across
all four sources); JPEG-$Q80$ post-degradation reduces all signals.}
\label{tab:pur-ablation-main}
\begin{tabular}{l|cccc|cccc}
\toprule
& \multicolumn{4}{c|}{Raw input} & \multicolumn{4}{c}{JPEG $Q80$} \\
Detector & Prod & AIg & FSw & Reen & Prod & AIg & FSw & Reen \\
\midrule
LP            & 0.91 & 0.78 & 0.77 & 0.78 & 0.71 & 0.62 & 0.56 & 0.58 \\
LoRA          & 0.53 & 0.66 & 0.64 & 0.61 & 0.44 & 0.56 & 0.53 & 0.51 \\
FullFT        & 0.61 & 0.75 & 0.76 & 0.78 & 0.44 & 0.62 & 0.64 & 0.69 \\
GenD-CLIP     & 0.65 & 0.65 & 0.66 & 0.65 & 0.65 & 0.60 & 0.63 & 0.62 \\
ForAda        & 0.41 & 0.50 & 0.66 & 0.55 & 0.44 & 0.48 & 0.65 & 0.54 \\
\textbf{EFFORT} & \textbf{0.98} & \textbf{0.81} & \textbf{0.89} & \textbf{0.89}
                & \textbf{0.79} & 0.58 & 0.68 & 0.72 \\
\bottomrule
\end{tabular}

\smallskip
\footnotesize
Bootstrap $95\%$ CIs (representative):
EFFORT $\times$ prod $0.982\,[0.961, 0.994]$;
EFFORT $\times$ cross-faceswap $0.891\,[0.883, 0.899]$;
EFFORT $\times$ cross-reenact $0.890\,[0.880, 0.898]$;
LP $\times$ prod $0.906\,[0.873, 0.937]$.
\end{table}

EFFORT is the only detector that generalizes broadly under raw input,
with AUROC $\geq 0.81$ on every source and $\geq 0.89$ on both
face-manipulation families. Linear probe comes second but plateaus
around $0.77$--$0.91$, leaving a clear gap on face manipulation.
JPEG-$Q80$ post-degradation costs every detector $0.15$--$0.25$
AUROC; EFFORT's face-swap AUROC drops from $0.89$ to $0.68$.

\paragraph{Multi-operator refinement and the JPEG kernel shift.}
We re-applied the multi-operator combination of
Section~\ref{subsubsec:pur-multi} --- now using raw output
$|\Delta\text{logit}|$ from four operators rather than internal
features --- to EFFORT individually. On raw input the multi-op LR
adds $+0.02$ to $+0.025$ AUROC on cross face-manipulation, lifting
cross-faceswap and cross-reenact each to $0.911$. On JPEG-$Q80$
input the multi-op LR adds $+0.04$ to $+0.09$ AUROC on production
($0.789 \to 0.878$) but fails to push cross face-manipulation above
the $0.85$ threshold ($0.681 \to 0.742$, $0.722 \to 0.761$).

\begin{table}[h]
\centering\small
\caption{Section~\ref{subsubsec:pur-ablation} --- multi-operator
logistic regression on four output $|\Delta\text{logit}|$ features
(median $3{\times}3$, median $5{\times}5$, Gaussian $\sigma{=}1$,
bit-depth $4$), applied to EFFORT. Under raw input the dominant
feature is median-$3$ ($+2.09$); under JPEG-$Q80$ it shifts to
median-$5$ ($+0.88$).}
\label{tab:pur-multiop-eff}
\begin{tabular}{l|cc|cc}
\toprule
& \multicolumn{2}{c|}{Raw input} & \multicolumn{2}{c}{JPEG $Q80$} \\
Subset & single med-3 & multi-op LR & single med-3 & multi-op LR \\
\midrule
prod              & 0.982 & 0.986 & 0.789 & \textbf{0.878} \\
cross-AI-gen      & 0.808 & 0.820 & 0.580 & 0.640 \\
cross-face-swap   & 0.891 & 0.911 & 0.681 & 0.742 \\
cross-reenact     & 0.890 & 0.911 & 0.722 & 0.761 \\
\midrule
\multicolumn{5}{l}{LR coefficient, dominant operator (others omitted):}\\
Raw input         & \multicolumn{2}{c|}{median-3 $= +2.09$} & & \\
JPEG-$Q80$ input  & & & \multicolumn{2}{c}{median-5 $= +0.88$} \\
\bottomrule
\end{tabular}
\end{table}

The coefficient shift between regimes is mechanistically meaningful.
Under raw input, the $3{\times}3$ median dominates ($+2.09$):
gradient-aligned adversarial perturbations live in the highest
spatial frequencies that this kernel removes. Under JPEG-$Q80$ input
the $3{\times}3$ median's contribution collapses to near zero
($-0.03$) and the $5{\times}5$ median takes over ($+0.88$). JPEG
quantization itself partially low-passes the input, so the
smaller-kernel response no longer carries adversarial-specific
information beyond what JPEG already removed; only the larger kernel
still pulls discriminative signal out of the JPEG-residual texture.
This is the structural reason to prefer multi-operator combinations
when input quality is heterogeneous.

\begin{figure}[t]
\centering
\includegraphics[width=0.85\textwidth]{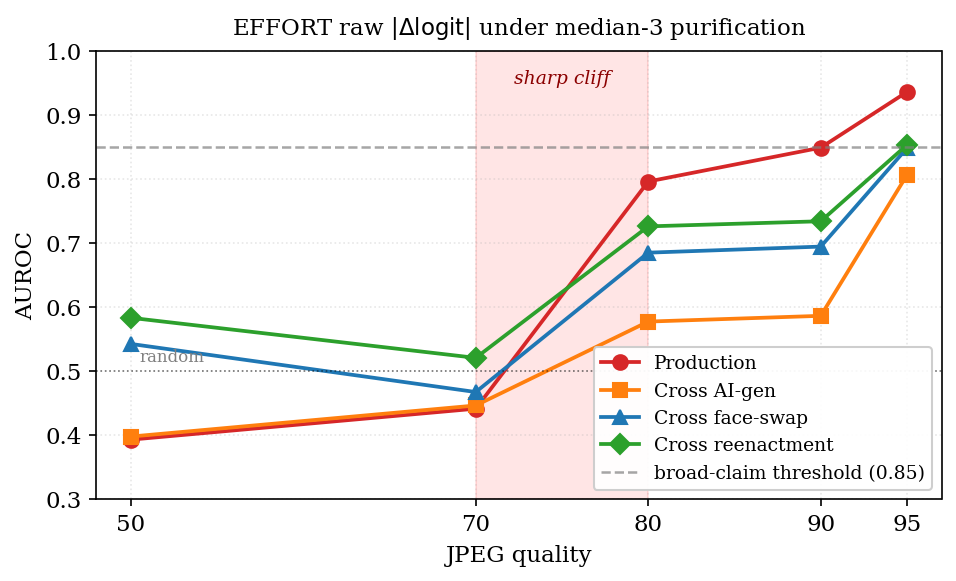}
\caption{Section~\ref{subsubsec:pur-ablation} --- EFFORT single
median-$3$ AUROC across the JPEG quality range. The sharp transition
between $Q70$ and $Q80$ (red shading) marks where adversarial
residuals survive compression sufficiently to be discriminated; at
$Q70$ and below all sources collapse to random. At $Q95$ the
single-operator signal alone clears the broad-claim threshold of
AUROC $0.85$ (dashed line) on every source: production $0.94$,
cross AI-gen $0.81$, cross face-swap $0.85$, cross reenactment
$0.85$.}
\label{fig:pur-jpeg-sweep}
\end{figure}

\paragraph{Sensitivity to JPEG quality.}
We swept JPEG quality from $Q50$ to $Q95$ with EFFORT and the single
median-$3$ operator (Figure~\ref{fig:pur-jpeg-sweep}). The
$Q70 \to Q80$ transition is a sharp cliff: cross-face-swap AUROC
jumps from $0.47$ (random) at $Q70$ to $0.69$ at $Q80$, and from
$0.69$ to $0.85$ between $Q80$ and $Q95$. At $Q95$ the single-operator
signal alone clears the broad cross-source claim (AUROC $\geq 0.85$
on every source). At $Q70$ and below, no operator combination tested
here recovers usable signal.

\paragraph{Diagnosis.}
On uncompressed input, raw $|\Delta\text{logit}|$ under EFFORT
separates adversarial from clean inputs across all four source types
(AUROC $0.81$--$0.98$, tight bootstrap CIs of representative width
$\pm 0.01$). This is the broadest cross-source generalization
observed in our exploration. The signal is detector-specific (the
other five same-backbone detectors plateau between $0.40$ and
$0.78$) and quality-dependent: the multi-operator LR partially
compensates at $Q80$, the single operator suffices at $Q95$, and no
combination we tested recovers usable signal at $Q70$ or below.

The original backbone-preservation hypothesis is not supported in
its strongest form: backbone parameter-update fraction alone does
not predict purification signal strength. Among our three controlled
trained detectors, the predicted ordering by decreasing backbone
preservation (LP $>$ LoRA $>$ FullFT) does not hold --- the actual
ordering is LP $>$ FullFT $>$ LoRA, with LoRA, which updates only
$\sim 1\%$ of parameters, the weakest of the three. Across all six
detectors EFFORT, which carries one of the heaviest backbone
modifications (SVD-residual decomposition over the full backbone),
is by far the strongest. The relevant property must therefore involve the specific subspace
targeted by the fine-tuning, or the smoothness of the resulting
input-output map. One plausible intuition is that SVD-residual
fine-tuning isolates the singular directions most strongly modified
by the deepfake-detection task, which may coincide with the
directions along which gradient-aligned adversarial perturbations
concentrate; removing those perturbations under purification would
then produce a disproportionately large output shift, while LoRA
and adapter modules operate in parameter subspaces less aligned
with the adversarial direction. We leave this conjecture untested
with the current experiments.

\paragraph{Implications.}
In practice, a self-defense pipeline based on this signal should
route purification scoring through EFFORT, switch to the
multi-operator combination when input quality is degraded, and fall
back to alternative signals below $Q70$. A mechanistic account of
why SVD-residual fine-tuning produces qualitatively different
purification behaviour, when the other detectors share its backbone
and several share its parameter-update fraction, remains open.

\subsection{Limitations and Future Work}
\label{subsec:purification-limitations}

\paragraph{Mechanism of EFFORT's generalization.}
Only EFFORT clears the cross-source $0.85$ AUROC threshold under raw
input; the other five same-backbone detectors plateau at $0.40$--$0.78$
and the ordering does not align with backbone-preservation fraction
(LP $>$ FullFT $>$ LoRA). Why SVD-residual fine-tuning produces this
qualitatively different response is not addressed by our data.
Promising probes are the input-output Jacobian smoothness per
detector, which singular directions each fine-tuning method modifies,
and controlled detectors with varied SVD ranks.

\paragraph{JPEG-$Q70$ cliff.}
No operator combination we tested recovers usable signal at $Q70$ or
below (cross-source AUROC $0.47$--$0.58$); heavy compression appears
to erase the high-frequency adversarial residual entirely.
Frequency-domain operators, perceptual-loss purification, and
calibration on quality-stratified validation sets are candidate
remedies that remain to be tested.

\paragraph{Single-detector dependency and unexploited layers.}
Routing all purification through EFFORT concentrates the failure
mode: an adversarial input crafted against EFFORT's gradient
geometry could plausibly evade both the classifier and the
purification signal. The output-only measurement that we used for
cross-detector comparison also leaves intermediate-layer signals
unexploited --- the multi-op result on GenD-DINOv3's $L_{19}$
(Section~\ref{subsubsec:pur-multi}) shows such layers can add
discriminative information, and a per-detector layer-sweep search
might recover usable signal on the five non-EFFORT detectors.

\paragraph{Attack coverage and self-defense framing.}
Our cross-distribution validation covers eight image sources with
PGD ($\epsilon \in \{2, 4, 8\}/255$), FGSM, and BIM targeting
ForAda, SigLIP+DINOv2, and a joint multi-detector objective. Novel
attack families (AutoAttack, transfer attacks from
foundation-model-scale detectors, attacks against unseen backbones)
are not covered, and the production result ($0.982$) is a
self-defense scenario in which our own pipeline generated the
adversarials. The gap between production and cross-source AUROCs
($0.98$ vs.\ $0.81$--$0.89$) is informative: detection degrades as
the attack distribution moves away from the training distribution,
and we expect realistic cross-attacker deployments to fall closer
to the cross-source numbers.

\section{Discussion}
\label{sec:discussion}

\subsection{Cross-Track Interpretation}
\label{subsec:discussion-crosstrack}

Viewed independently, our generation, detection, and further-experiments
tracks report separate outcomes; viewed together they trace a single
underlying tension. Section~\ref{sec:generation} demonstrates that a
moderately sophisticated adversarial pipeline can drive the fake-class
probability of most off-the-shelf detectors below their decision
threshold within an $\epsilon = 2/255$ budget.
Section~\ref{sec:detection} implicitly encounters the consequence: any
classifier operating on raw pixels --- regardless of its representational
backbone --- is exposed to this manipulation when adversarially
post-processed inputs appear in the wild. Section~\ref{sec:purification}
reframes the problem as a two-stage defence: rather than continuing to
raise classification accuracy against arbitrarily adaptive adversaries,
it asks whether the same detectors can serve as a second-order signal
that \emph{flags} the presence of adversarial post-processing, and
identifies EFFORT under median-3 purification as a working
instantiation. Read as a whole, the three tracks are not three
independent contributions but three moves in one attacker--defender
exchange, with the purification signal emerging as a natural
architectural complement to --- rather than a replacement for --- the
underlying classifier.

\subsection{Strengths of the Proposed Approach}
\label{subsec:discussion-strengths}

Each track exhibits distinct strengths. On the generation side, the
iterative composition of DiffJPEG-in-loop, MI/DI/EoT transferability
techniques, and two-stage warm-start achieved $\geq 95\%$ evasion on
15 of 17 evaluated detectors (Table~\ref{tab:evasion}) and $0.9019$
on the held-out organizer detector pool, demonstrating that the
attack ceiling under $\epsilon = 2/255$ lies well above what standard
PGD alone reaches. On the detection side, the complementary
specialists ensemble (SigLIP+DINOv2 $\times$ GenD-DINOv3) achieved
$0.9939$ baseline-deepfake and $0.8822$ participant-deepfake accuracy
--- outperforming any single unified detector we surveyed
(Table~\ref{tab:model_comparison}) --- with the design empirically
justified by demonstrating that no single unified detector covered
both fake regimes simultaneously in our benchmark. On the purification
side, three distinct contributions emerge from the investigation.
Empirically, the fine-tuning ablation across six same-backbone
detectors establishes fine-tuning strategy --- rather than backbone
identity or update fraction --- as the variable that governs
purification signal strength, with EFFORT's SVD-residual decomposition
yielding AUROC $0.81$--$0.98$ across four adversarial source types.
Theoretically, this finding refutes the simple backbone-preservation
hypothesis: LoRA, which updates only $\sim 1\%$ of parameters, produced
the weakest signal, while EFFORT with substantially heavier backbone
modification produced the strongest. Methodologically, the raw
$|\Delta\text{logit}|$ signal requires no per-detector or
per-distribution recalibration --- unlike the multi-layer Mahalanobis
ensemble of Section~\ref{subsec:maha} that collapsed at deployment ---
making it operationally attractive as a first-pass adversarial screen.

\subsection{Limitations}
\label{subsec:discussion-limitations}

Limitations similarly cut across tracks. On the generation side, the
transferability ceiling to unknown participant architectures ($0.5762$
vs. $0.9019$ on organizers) is a structural rather than tuning-level
constraint, and face-crop evasion remains infeasible within the
$\epsilon = 2/255$ budget (Challenge~6,
Section~\ref{subsec:challenges}). On the detection side, our
count-based $\tau_{\min}$ optimisation against a Real:Fake-imbalanced
calibration set introduced a compounding pair of errors
(Section~\ref{subsec:det-results}) that ROC-based per-detector
threshold selection would have avoided. On the purification side, four open issues bound the result:
the unresolved mechanism behind EFFORT's cross-source specificity,
the JPEG-$Q70$ signal cliff that erases the high-frequency
adversarial residual, single-detector routing as a bottleneck
against adaptive attackers, and attack-family coverage restricted
to PGD/FGSM/BIM. Detailed treatment of each appears in
Section~\ref{subsec:purification-limitations}. Collectively, the
generation and detection limitations point to concrete engineering
remedies (broader transfer objectives, rate-based calibration),
while the purification limitations invite deeper mechanistic study.

\subsection{Lessons Learned}
\label{subsec:discussion-lessons}

Three methodological lessons emerge from the purification investigation
and transfer to any future work in this space. First, the JPEG-quality
confound is structural rather than incidental --- it defeated all
three adversarial-signal families in Section~\ref{sec:purification},
motivating quality-stratified validation in any future deployment.
Second, pilot-scale AUROC systematically overstates deployment-scale
performance for any signal whose decision boundary depends on
calibration-data statistics, with the multi-layer Mahalanobis
ensemble's drop from CV AUROC $1.0$ to test-set $0\%$
(Section~\ref{subsec:maha}) the most extreme instance we observed.
Third, sigmoid saturation can conceal substantial raw-logit signals
from probability-domain metrics; we initially missed a $36\times$
linear-probe separation this way, and raw logit shifts must be
checked before concluding that a signal has failed. Two additional
strategic lessons emerge from the competition side: rate-based rather
than count-based metrics should govern any threshold calibration
against distribution-shifted test sets
(Section~\ref{subsec:det-results}), and held-out cross-architecture
evaluation on the attack side is essential to estimate the
transferability gap before submission
(Section~\ref{subsec:gen-results}).

\section{Conclusion}
We presented our approaches to the two official sub-tasks of the
ImageCLEF 2026 Deepfake Detection and Generation Task together with
a self-initiated follow-up investigation. Our generation pipeline
achieved a final score of 0.4170; our detection ensemble achieved
0.6986. Beyond the competition, we investigated purification-based
adversarial detection across six detectors sharing a CLIP ViT-L/14
backbone, identifying EFFORT under median-3 purification as the
broadest cross-source generalizer (AUROC 0.81--0.98 across four
adversarial source types) --- a finding that refutes the simple
backbone-preservation hypothesis. A mechanistic account of why
SVD-residual fine-tuning produces qualitatively distinct
purification behaviour, and remedies for the JPEG-$Q70$ signal
cliff, remain open directions for future work.

\section*{Author contributions}
J. Kim conceived the methodology, designed and conducted all
experiments (image generation pipeline, detection ensemble, and the
purification-based adversarial detection investigation), and authored
the manuscript. S. Kim provided cross-track discussion and manuscript review. J. Woo provided supervisory
feedback on the manuscript.

\begin{acknowledgments}
This research was supported by the Ministry of Science and ICT (MSIT), 
Korea and the Institute of Information \& Communications Technology 
Planning \& Evaluation (IITP) under AI University (2026-0-00032, 2026).
\end{acknowledgments}

\section*{Declaration on Generative AI}
During the preparation of this work, the author(s) used Claude
(Anthropic) in order to: Drafting content, Grammar and spelling
check, Paraphrase and reword, Improve writing style, Peer review
simulation. After using these tool(s)/service(s), the author(s)
reviewed and edited the content as needed and take(s) full
responsibility for the publication's content.

\bibliography{references}

\end{document}